\documentclass[10pt,twocolumn,letterpaper]{article}

\usepackage[pagenumbers]{cvpr} 

\usepackage{graphicx}
\usepackage{amsmath}
\usepackage{amssymb}
\usepackage{booktabs}
\usepackage[nolist]{acronym}

\usepackage[pagebackref,breaklinks,colorlinks]{hyperref}

\usepackage[capitalize]{cleveref}
\crefname{section}{Sec.}{Secs.}
\Crefname{section}{Section}{Sections}
\Crefname{table}{Table}{Tables}
\crefname{table}{Tab.}{Tabs.}

\def\cvprPaperID{11662} 
\def\confName{CVPR}
\def\confYear{2022}

\newcommand{\eal}[1] {#1 et al.}

\begin{document}

\title{METEOR Guided Divergence for Video Captioning}

\author{Daniel Lukas Rothenpieler\\
Chair of EIHW\\
University of Augsburg, Germany\\
{\tt\small rothenpielerdaniel@gmail.com}
\and
Shahin Amiriparian\\
Chair of EIHW\\
University of Augsburg, Germany\\
{\tt\small 	shahin.amiriparian@uni-a.de}
}
\maketitle

\begin{abstract}
Automatic video captioning aims for a holistic visual scene understanding.
It requires a mechanism for capturing temporal context in video frames and the ability to comprehend the actions and associations of objects in a given timeframe.  Such a system should additionally learn to abstract video sequences into sensible representations as well as to generate natural written language. While the majority of captioning models focus solely on the visual inputs, little attention has been paid to the audiovisual modality.
To tackle this issue, we propose a novel two-fold approach. First, we implement a reward-guided KL Divergence to train a video captioning model which is resilient towards token permutations. Second, we utilise a \ac{BMHRL} Transformer architecture to capture long-term temporal dependencies of the input data as a foundation for our hierarchical captioning module.
Using our \ac{BMHRL}, we show the suitability of the \acs{HRL} agent in the generation of content-complete and grammatically sound sentences by achieving $4.91$, $2.23$, and $10.80$ in \acs{BLEU}3, \acs{BLEU}4, and \acs{METEOR} scores, respectively on the \textit{\mbox{ActivityNet} Captions} dataset.
Finally, we make our \ac{BMHRL} framework and trained models publicly available for
users and developers at \url{https://github.com/d-rothen/bmhrl}.
\end{abstract}

\section{Introduction}
\label{sec:intro}

Video Captioning is the task of automatically generating contextual descriptions of videos.
It incorporates computer vision alongside \ac{NLP}, and generally employs (deep) \ac{CNN}s to represent a video's content in a defined feature space and use said features in a \ac{RNN} architecture to generate textual descriptions\cite{seq2seq-cap,bidir-rl-captioning, deep-rnn-cap}.

The dawn of Transformers has been central in advancing sequence-to-sequence models, becoming an important architecture not only for \ac{NLP} but also vision and audio~\cite{visual-transformers, speech-transformer}. A broad amount of multimodal training data~\cite{krishna2017densecap} has become usable with the increase in bimodal architectures~\cite{BMT_Iashin_2020}. The increasing complexity of high-resolution audio and video recordings has constrained a majority of video captioning models to employ offline trained feature extractors~\cite{I3D, VGGish}, thus benefiting from the extensive representation learning of prior work.
However, multiple end-to-end Transformers-based approaches, \eg \textsc{SwinBERT}~\cite{SwinBERT}, or the end-to-end generative pretraining system proposed by \eal{Seo}~\cite{end-to-end-pretraining}, which take video frames  as raw input and output natural language descriptions have shown to be effective for the task of multimodal video captioning.
While a lot of work is focusing on realising rich video feature generation \cite{chen-motion-guided-cap,BMT_Iashin_2020,bidir-temp-graph-cap,pan-spatio-temp-graph-cap} fewer endeavours consider the modelling of objective functions and their implications, where linguistic properties can be exploited to accurately optimise language models towards robust sentence generation\cite{hier-modular-cap,Wang_HRL}.
\ac{RL} has been found to be an adequate strategy for accurate sentence generation, as strict word-by-word-based learning leads to outputs suffering from exposure bias as well as loss-evaluation mismatch~\cite{RanzatoCAZ15}. Furthermore, \ac{HRL} architectures~\cite{move_forward_and_tell_xiong2018, StreamlinedDenseCap} could make use of the temporally-dependent context of subsequent events and captions in video captioning.
Opting to make use of strongly defined sentence structure, \eal{Wang}~\cite{Wang_HRL} designed a hierarchical architecture to guide sentence generation towards grammatically coherent synopses.\\\\
In this study, we seek to combine different advancements in sequence-to-sequence modelling in building a bimodal Transformer architecture with a \ac{HRL} module.
We use the ActivityNet Captions dataset~\cite{krishna2017densecap} to train and validate our model.
The architecture builds on top Iashin and Rahtu's~\cite{BMT_Iashin_2020} work which is available at \url{https://github.com/v-iashin/BMT}. The \ac{HRL} methodology is inspired by \eal{Wang}~\cite{Wang_HRL}.
We use I3D~\cite{I3D} and VGGish~\cite{VGGish} to extract video and audio features from the source material, respectively. \\
\\
Our main contributions are twofold:\\
First, we introduce a biased \ac{KL} Divergence that is guided by a two-staged hierarchical \ac{METEOR} score facilitating a  resilient model towards token permutations. By doing so, changes in the word order have less impact on penalising the loss values.\\
Second, we present the fusion of bimodal Transformers decoded representations in a \ac{HRL} module. Thereby, we aim for capturing long(er) temporal context (compared to recurrent networks) from the input features, building a solid basis for the \ac{HRL} module to enhance the overall model performance.
\section{Related Works}
\label{sec:related}

\textbf{Sequence-to-sequence modelling} has seen significant improvements with the prevalence of Transformer architectures which have shown to generally outperform the capabilities of \ac{RNN} architectures for \ac{NLP}~\cite{vaswani2017attention}.
While visual representations tend to be of much higher dimensionality than their text-based counterparts, similar architectures have been used for computer vision, either directly on visual inputs~\cite{dosovitskiy2020vit} or latent visual features\cite{masked-trafo-cap, BMT_Iashin_2020}.
Iashin and Rahtu \cite{BMT_Iashin_2020} used audiovisual representations in a crossmodal Transformer architecture, leveraging long-range contingencies for sentence generation and making use of the additional information provided by the audio modality. 
As video captioning mostly disregards real-time processing, several approaches have made use of exploiting both past and future video context to further improve scene understanding~\cite{bidir-att-fus-cap,bidir-temp-graph-cap}.\\

\textbf{Reinforcement learning} has been successful in circumventing different problems posed by traditional supervised methods. 
Captioning models tend to be fully supervised on the videos labelled with captions, using divergence metrics to align the model's output with the \ac{GT} distributions. This in turn leads to an objective mismatch\cite{RanzatoCAZ15}, where models are trained to minimise the divergence between output and \ac{GT} words while the general performance is most commonly measured by machine translation metrics such as \ac{BLEU}~\cite{papineni-etal-2002-bleu}, \ac{CIDEr}~\cite{cider-score} and \ac{METEOR}~\cite{banerjee-lavie-2005-meteor} which are not differentiable. In order to be aligned with these metrics, REINFORCE based approaches~\cite{RanzatoCAZ15,entailment-reward-cap,Wang_HRL} went on to employ policy gradients to maximise expected scores over predicted sequences.\\

\textbf{Inherent hierarchical structures} are an integral part of video captioning.
First and foremost sentence structure plays a key role~\cite{Wang_HRL, syntax_cap} when attempting to capture a scene's subjects and predicates.
Sentences can be broken down into individual clauses, which can be used in supervision for partial sentence generation.
Wang \etal~\cite{Wang_HRL} use \ac{HRL} to train a manager module which defines abstract sentence goals for a low-level module to complete, guiding sentences to follow the learned sentence structure.
Aggregated \ac{BLEU} scores are used to encourage correct clause generation.

Furthermore, video captioning follows the general paradigm of visual cognition with subsequent language generation. While long sequences are broken down into individual events, (visual) features can provide context beyond their immediate appearance in a video's segment\cite{hrnn-paragraph-cap, StreamlinedDenseCap}.
Mun \etal~\cite{StreamlinedDenseCap} utilise visual context generated from their event proposal module to condition captions per event.
Their captioning module uses \ac{HRL} rewards to measure outputs both at the event and clip levels.
Recently, Ye \etal~\cite{hier-modular-cap} employed a \ac{DETR} inspired approach to detect objects as latent features for the downstream sentence generation, having two attention modules predict the \ac{GT} predicate and then the full sentence conditioned on detected objects where all modules are supervised individually with the corresponding linguistic embeddings using SBERT\cite{reimers-sbert}.

\section{Methodology}
\label{sec:methodology}

 \begin{figure*}[h!]
      \centering
        \includegraphics[width=.8\textwidth]{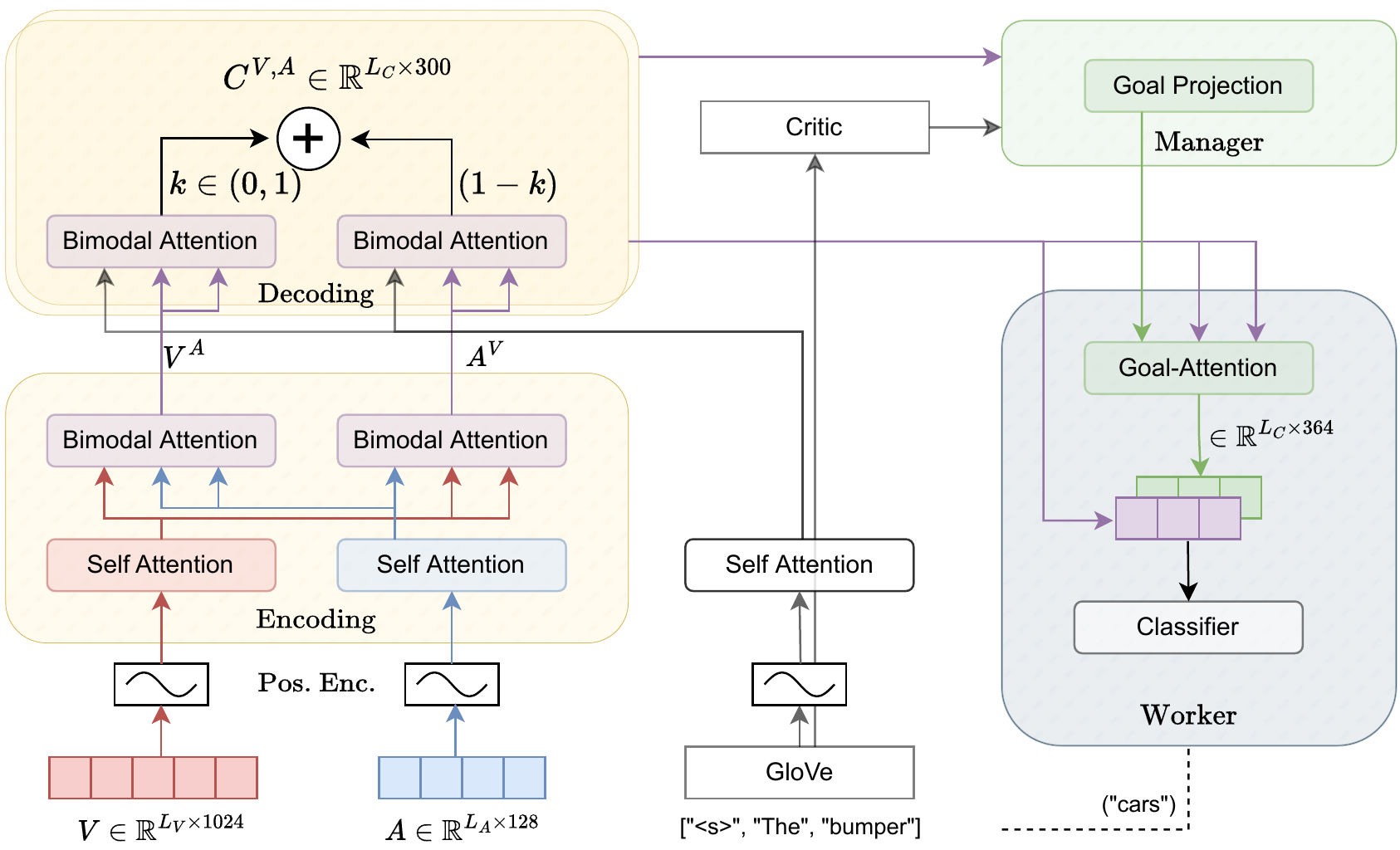}
      \caption{\ac{BMHRL} architecture. Both \ac{GloVe} and Critic modules are pretrained. All input features are positionally encoded to maintain temporal information throughout the attention layers.}
\label{fig:model}
\end{figure*}

    We propose a bimodal architecture, denoted as \ac{BMHRL} (cf.~\Cref{fig:model}) composed of two main components:
    (i) A bimodal encoder-decoder (cf.~\Cref{sec:features}) architecture with (ii) a \ac{HRL} module stacked on top (cf.~\Cref{sec:hier-rew}).
    Via the first component, audio and video representations bimodally attend to one another and then further attend to the \ac{GloVe} embeddings of previous iterations.        
    The Manager module from the second component, tasked with guiding a subservient Worker module towards the completion of sentence clauses, generates a goal vector. When the Critic module, trained on the Charades Caption dataset\footnote{Charades Caption: \url{http://www.cs.ucsb.edu/~xwang/data/CharadesCaptions.zip}}~\cite{Wang_HRL} specifies a sentence clause to be complete, a new goal-vector will be determined.
    The Worker attends to the generated goal and classifies the next word. Audio ($A$) features are represented by $128$-d extracted VGGish~\cite{VGGish} features. Video ($V$) features 
    are extracted via I3D~\cite{I3D} into $1024$-d feature vectors. A single feature vector represents $.96$ and $2.56$ seconds of source material, respectively.
    Additionally, we feed the network $300$-d \ac{GloVe} embeddings of the previously generated captions in order to provide further context.
    We utilise \ac{METEOR}-based rewards alongside \ac{KL} Divergence to train our model. 

\subsection{Features}
\label{sec:features}
The proposed model uses I3D~\cite{I3D} features, pretrained on the Kinetics dataset~\cite{kinetics-dataset} computed over $64$ RGB and optical flow frames with a sampling rate of $25$\,fps and $224\times224$ resolution. For audio features VGGish~\cite{VGGish}, trained on AudioSet~\cite{gemmeke2017audio}, was used to generate log Mel spectrogram representations over a $.96$ second window. 
The total dataset~\cite{krishna2017densecap} contains annotations for 20k YouTube videos with more than 100k temporally localised sentences, of which $50\%$ was used in training~\cite{BMT_Iashin_2020}.\\\\
Both $V$ and $A$ features are first passed through a self-attention layer, using four attention heads, to obtain $V^V$ and $A^A$. The features are further passed through the bimodal attention mechanism, generating the audio-attended visual features $V^A \in \mathbb{R}^{L_V\times d_L}$ and visual-attended audio features $A^V \in \mathbb{R}^{L_A\times d_L}$.

    \begin{figure}[htb]
          \centering
          \includegraphics[width=1\columnwidth]{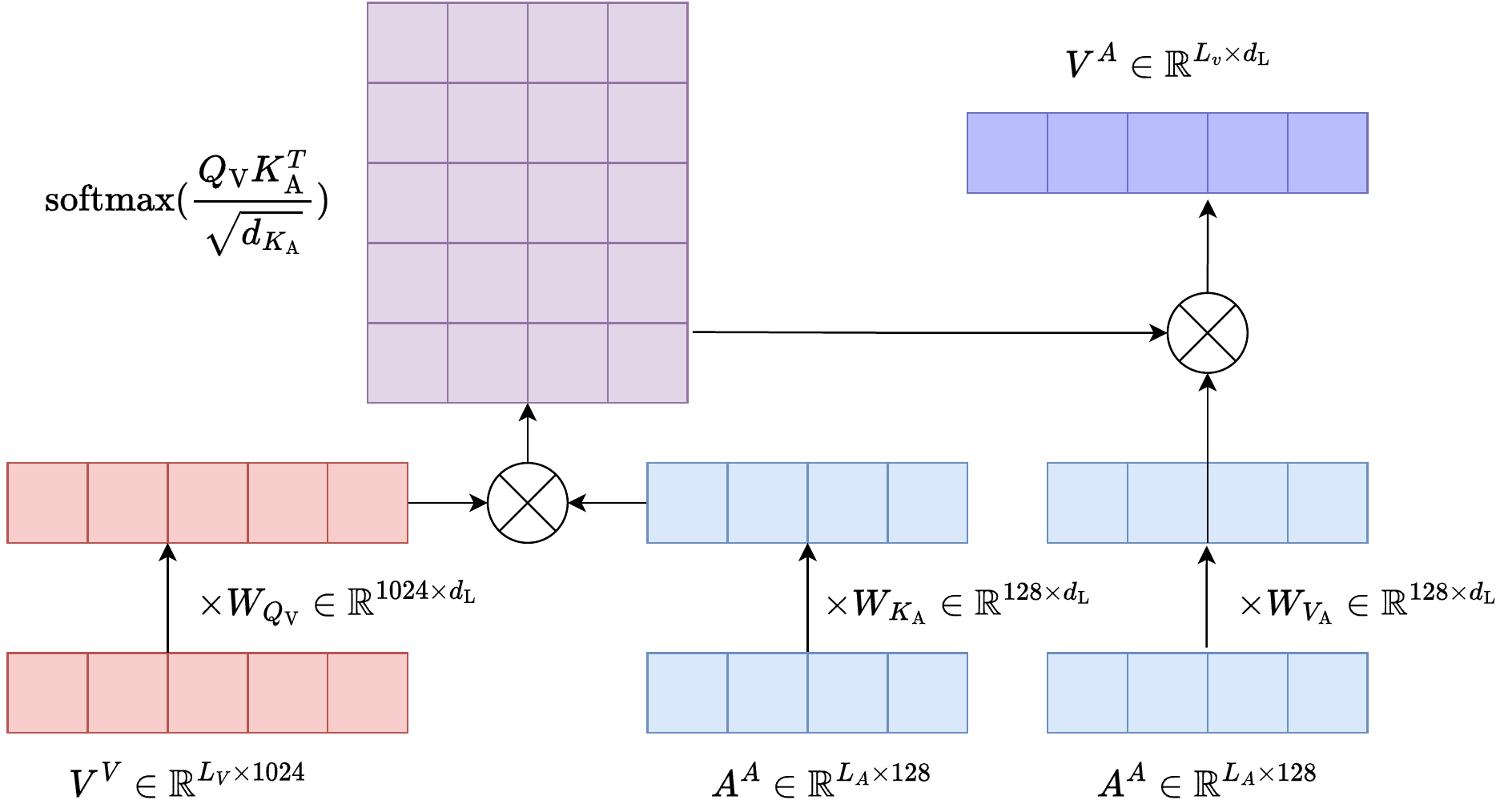}
          \caption{Bimodal Attention (BMA). We use a dimension of $d_L = 1024$ for the latent feature space.}
    \label{fig:bma}
    \end{figure}

\subsection{Decoding}
\label{ssec:decoding}
	As shown in~\Cref{fig:bma}, we make use of two identical decoder blocks.
	The Worker's designated decoder is tasked with producing low-level features that are used in the word-level classification step. In contrast, the Manager's decoder can focus on high-level abstraction to contextualise sentence clause generation.

	Each block is fed with the bimodal representations $V^A, A^V$, as well as the self-attended target caption $C$~\cite{BMT_Iashin_2020}.
	We compute the bimodal attention for both $V^A$ and $A^V$ utilising a similar bimodal attention mechanism as the encoders' attention layers.
	The resulting caption-attended features $C^{V^A}$, $C^{A^V}$ are then weighted with a learnt constant $k$, squashed via the Sigmoid function ($\sigma$) to generate a weight factor between $0$ and $1$.
	This feature fusion is then defined by
	\begin{equation}
	    C^{V,A} = \sigma{(k)} * C^{V^A} + (1 - \sigma{(k)}) * C^{A^V}.
	\label{eq:adaptive_fusion}
	\end{equation}
	
	As such, we allow the model to put emphasis on the representations of higher importance as well as generating a shared feature space ($\in \mathbb{R}^{d_L}$) of the bimodal representations.

\subsection{Goal Generation}
\label{ssec:goal-generation}
    When training the Manager module, the model, apart from the Manager's decoder and itself, is frozen in order to hinder inference from learning a hierarchical structure in parallel.
    The Manager's task is to guide the Worker towards generating a segment $s_t = (\hat{y}_t, \hat{y}_{t+1}, ..., \hat{y}_{t+i})$
    until in the $t+i$-th timestep, the Critic considers the generated segment as fulfilled, at which point the Manager will output a new goal~\cite{Wang_HRL}.
    We linearly transform the \ac{HRL} features $C^{V,A}$ to shape $L_C\times d_{\text{goal}}$.
    This module is treated as a deterministic \ac{RL} policy. We introduce noise in order to allow for \emph{action exploration}.
    We choose a normal distribution with adaptive mean and standard deviation relative to the goal vector's values.
    Features that do not align with the Critic's determined segment boundaries are dropped.
    The goals' feature dimension is $d_{\text{goal} = 64}$.

\subsection{Classification}
When training the Worker module, the Manager as well as its decoder are frozen. The Worker serves as the classifying module of the \ac{HRL} agent. Using the bimodal attended \ac{HRL} features $C^{V,A}$ as key and value, the generated features attend to the Manager's goal. By doing so, we quantify the progress of the Worker's goal completion. The goal-attention vector is then concatenated with the \ac{HRL} features, resulting in the classification input $\hat{C}^{V,A}_{g} \in L_C\times (d_C + d_{\text{goal}})$. Finally, we use a linear layer to project the appended representation into the vocabulary space of $d_\text{vocab} = 10\,172$.

\subsection{Hierarchical Rewards}
\label{sec:hier-rew}
We distinguish between word-level (Worker) $R_W$ and segment-level (Manager) $R_M$ reward. The Worker's score $\Delta\text{METEOR}_W$ is implemented as an equivalent to the proposed reward by Wang et al.~\cite{Wang_HRL}. Sampling words from the Worker $\hat{y} \sim \pi_w$, we obtain $\hat{Y} = (\hat{y}_0, \hat{y}_1, ..., \hat{y}_n, ..., \hat{y}_m)$. We compute the incremental \ac{METEOR} score\footnote{Used package: \url{https://www.nltk.org/api/nltk.translate.meteor_score.html}} of each timestep $t$'s partial predicted output $(\hat{y}_0, ..., \hat{y}_t), t\leq m$ with the \ac{GT} sentence $Y = (y_0, y_1, ..., y_n)$.

\begin{equation}
    	\begin{aligned}
    \Delta\text{METEOR}(\hat{Y}, Y, 0) &= \text{meteor}((\hat{y}_0), Y)\\
	\Delta\text{METEOR}(\hat{Y}, Y, t) &= \text{meteor}((\hat{y}_0, \dots, \hat{y}_t), Y) - \\
	&\text{meteor}((\hat{y}_0, \dots, \hat{y}_{t-1}), Y)
	\end{aligned}
\end{equation}

    	Rewards are then added up and discounted with factor $\gamma$, such that
	\begin{align}
    R_W(t) = \sum_{j=t}^n \Delta\text{METEOR}(\hat{Y}, Y, j) * \gamma^{j-t}
	\end{align}

	With the Worker's word-level rewards fully computed, we can calculate the Manager's reward by summing over the Critic's segment boundaries. For a single segment $s_k = (y_k, ..., y_n)$, we define the high-level reward as
	
\begin{equation}
    \begin{aligned}
                R_M(k, n) = \sum_{j=k}^{k+n} \Delta\text{METEOR}(\hat{Y}, Y,j)\\+ R_M(k+n+1, m),
    \end{aligned}
\end{equation}
	where $m$ is the length of the next segment. If $k$ does not denote a segment's first position, then $R_M(k,\cdot) = 0$

\subsection{Signal Function}
\label{sec:signal_f}
\begin{figure}[htb]
          \centering
          \includegraphics[width=1\columnwidth]{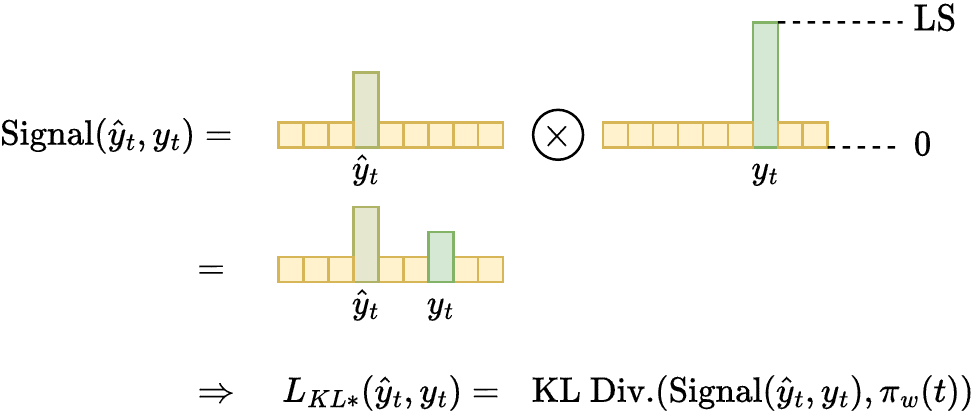}
          \caption{Combining the model prediction $\hat{y}_t \sim \pi_w$, with the \ac{GT} $y_t$.
          Instead of fully relying on the \ac{GT} signal, we add the scaled reward of the current prediction while removing its value from the \ac{GT} signal.}
    \label{fig:kl-signal}
    \end{figure}

We construct a signal function (cf.~\Cref{fig:kl-signal}) utilising the $\Delta\text{METEOR}$ and \ac{KL} Divergence with the \ac{GT} signal. We form a single spike distribution
over the \ac{GT} words $d^y_{LS}$ with $d^y_{LS}(y) = LS$ and $d^y_{LS}(w) = \frac{1-LS}{d_{\text{vocab}}-2}, w \neq y$, $LS$ being a \textit{Label Smoothing} constant of $.3$.

The sampled probabilities $\pi_w (\hat{y})$ are scaled with the modified reward and the synopsis' length.
Using a baseline function $b(\hat{C}^{V,A}_{G})$, trained in parallel to the model, and reward $R(t) \in \{R_W(t), R_M(t,\cdot)\}$ at time $t$, we create an advantage factor $\eta_t$, balancing the target signal between the \ac{GT} Logit and sampled prediction $\hat{y}_t$.
We denote the reward-scaled target distribution as $d^{y}_{RS}$ based on the following equations:
\begin{equation}
	    \eta_t = (R(t) - b_t) * L_C * \pi_w(\hat{y})\\
\end{equation}
\begin{equation}
    d^y_{RS}(y) = (1-\eta_t)  d^y_{LS}(y)
\end{equation}
\begin{equation}
    d^y_{RS}(\hat{y}) = \eta_t (1-c_{\text{smooth}})
\end{equation}

 For each word $w \not\in \{\hat{y}, y\}, d^y_{RS}(w) = d^{y}_{LS}(w)$.\\
		
    When $y = \hat{y}$, $d^y_{RS}$ will be equal to $d^y_{LS}$. Thus, we increase the robustness of the model with respect to short-sighted optimisation as we focus on a prediction's effect on the whole following sentence instead of its immediate divergence value.
    We task the model to learn trajectories such that generated sequences have minimal compounding error.
    
\begin{figure*}[h!]
    \centering
        \includegraphics[width=.8\textwidth]{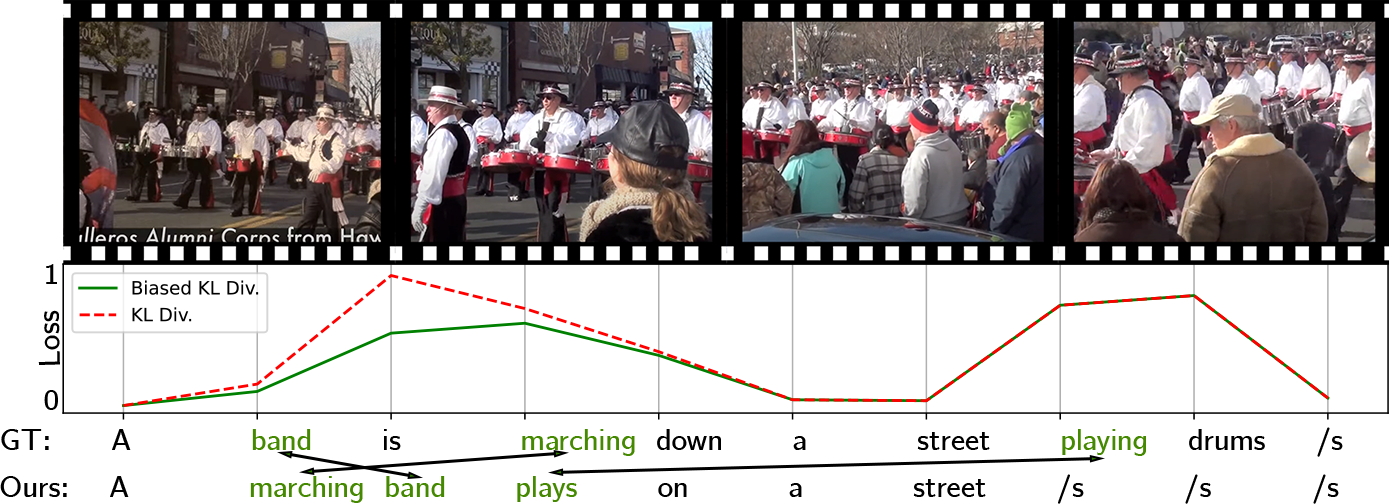}
     \caption{Comparison of the resulting \ac{KL} Divergence using our proposed biased \ac{KL} Divergence against the standard divergence. Words that positively influence the (normed) loss value through their relationship with the \ac{GT} are connected with arrows. Here we observe how similar words in differing order lead to smaller divergence. Sampled words were taken greedily from the predicted distribution to compute the amplitude. }
    \label{fig:biased-kl-result-a}
    \end{figure*}

\begin{figure*}[h!]
    \centering
        \includegraphics[width=.8\textwidth]{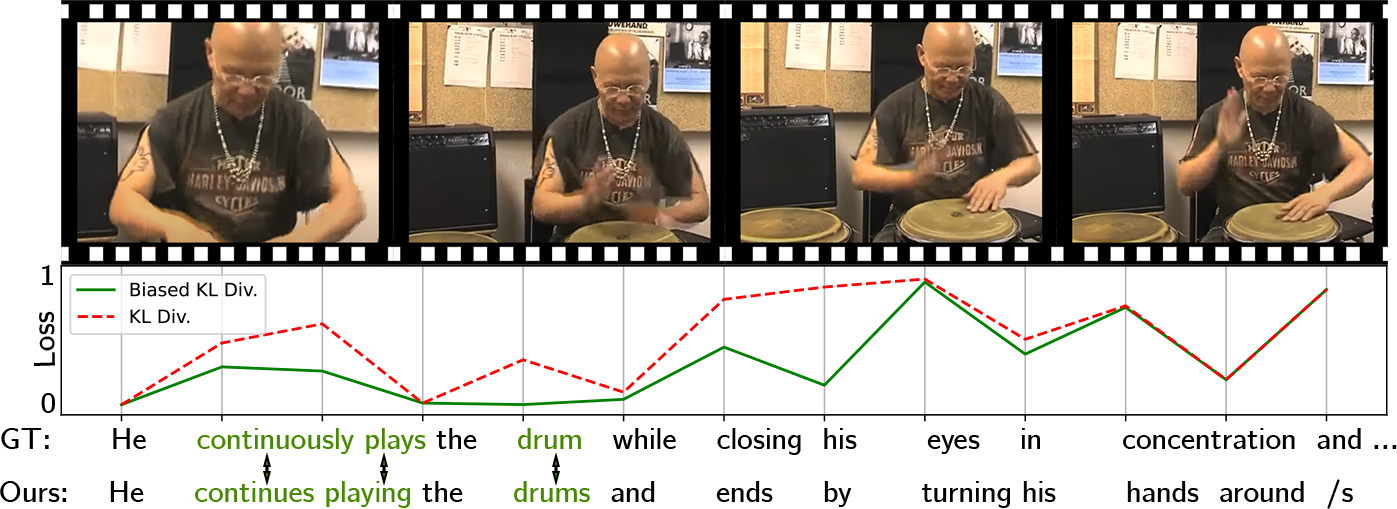}
    \caption{ This example highlights the tolerance of our model with regard to similar word stems with the \ac{GT}. Albeit being in matching positions, a standard divergence metric cannot account for related words.}
    \label{fig:biased-kl-result-b}
    \end{figure*}

    Subtle permutations of predicted tokens are resilient to changes in reward, whilst resulting in high \ac{KL} divergence (cf.~\Cref{fig:biased-kl-result-a,fig:biased-kl-result-b}).

\section{Experiments}
\label{sec:experiments}

In the process of evaluation, we tuned several hyperparameters and performed an ablation study.
The \ac{BMHRL} model itself is trained first as a bimodal Transformer (cf.~\Cref{sec:abl_study}) and gets further fine-tuned with the \ac{HRL} setting.

    \subsection{BMHRL}
    The \ac{BMHRL} uses a latent feature representation of $1024$, and $1024, 256, 300$ for the vision, audio, and text modalities, respectively.
    We use discount factors of $.7$, and $.8$ for the Worker and Manager rewards.

    \noindent\textbf{Weighted \ac{BMHRL}:} We alternatively used the $\eta_t$ advantage factor (cf.~\Cref{sec:signal_f}) as an inverse scaling value for the models \ac{KL} divergence with $d_{LS}$.
    For this experiment, $\eta_t$ was clamped between $(0,1)$ and further scaled with a normalisation constant.\\
    \noindent\textbf{Dimensionality adjustment:}
    With this model iteration, we changed the modalities' feature representations to $512$ and $1024$ for audio and text. The number of attention layers in both the encoder and decoder modules was increased from 2 to 3.
    \\
    \noindent\textbf{Discount factor adjustment:} We increase the discount factor of both manager and worker to $.8$ and $.9$.
    This way, we hope to observe how the reliance on later predictions affects total loss and model performance.
    Similarly, in order to account for employing teacher forcing alongside discounted \ac{METEOR} rewards, we set the discount factors to $0$.
    \subsection{Ablation Study}
    \label{sec:abl_study}
    In order to control for the effects of combining hierarchical rewards and different modalities, we omit several parts of our model in the following experiments.\\
    \textbf{BMH}: 
    This approach uses the \ac{BMHRL} architecture without utilising the $\Delta\text{METEOR}$ rewards to manipulate the Label Smoothing signal. It works similarly to the \ac{BMT}.
    \\
    \noindent\textbf{Audio Only:}
    Here, we omit the visual modality and bimodal encoding step. The self-attended audio representations are bimodally decoded with the \ac{GloVe} embedded captions and further processed in the models \ac{HRL} head. 
    
    \noindent\textbf{Vision Only:}
    This experiment was done analogously to the audio-only approach, using visual features instead.

    \section{Results}
    \label{sec:results}


    \textbf{BMHRL:}
    Our model achieves $4.91$, $2.23$ and $10.80$ in \ac{BLEU}3,4 and \ac{METEOR} scores, respectively (cf.~\Cref{tab:model_performances}).
    We observe an absolute increase of $.28$ points in \ac{BLEU}4, $.3$ in \ac{BLEU}3 and $.10$ reduction in \ac{METEOR} scores upon the BMT~\cite{BMT_Iashin_2020}, on which our model's architecture is based on.
    \eal{Krishna}~\cite{krishna2017densecap}'s model follows the labelling structure more closely, thus having substantial performance benefits in \ac{BLEU}3,4 scores while the \ac{BMHRL} manages to capture video contents to a greater extent, resulting in high \ac{BLEU} and 
    \ac{METEOR} scores.
    A recurrent network, optimised with hierarchical rewards, by \eal{Mun}~\cite{StreamlinedDenseCap}, manage to further increase the \ac{METEOR} score by $2.27$.
    In this case, our model is less descriptive  while forming improved coherent sentences with respect to the test set's labels.
    The vocabulary usage increases on our baseline by 143 words. Vocabulary size indicates the expressiveness of the trained models. For the audio only model, its comparatively high \ac{METEOR} score in combination with a low vocabulary usage shows the model's inclination to fall back to the highest frequency words present in the dataset.
Unfortunately, other mentioned architectures did not publish this metric.
    \\
    \noindent\textbf{BMH:}
    The model takes about twice as many training episodes to converge while performance scores \ac{BLEU} and \ac{METEOR} are slightly diminished.
    Differences with the BMT scores emerge due to individual architectural design choices.
    \\
    \noindent\textbf{Audio only:}
    The audio-only approach still reaches surprisingly competitive machine translation scores, however, its vocabulary usage is mostly constrained to the most frequent words in the dataset's labels.\\
    \noindent\textbf{Vision Only:}
    As expected, the vision modality naturally outperforms the audio-only modality but falls behind the BMH. 

    \begin{table}[ht]

    \centering
    \resizebox{1\columnwidth}{!}{    
    \begin{tabular}{lrrrrr}
         \toprule
         & & \multicolumn{3}{c}{Performance Metrics} & \\
         \cmidrule(lr){3-5} 
         Model & RL & BLEU3 & BLEU4 & METEOR & Vocab. size\\
         \midrule
         \multicolumn{6}{l}{Related Works}  \\
         \midrule
         BMT~\cite{BMT_Iashin_2020}& - &  4.61  & 1.95   &10.90 & 944\\
         Krishna et al.~\cite{krishna2017densecap} & - &\textbf{7.12} &\textbf{3.89} &  9.46&?\\
         Mun et al.~\cite{StreamlinedDenseCap}& \checkmark &   4.41  & 1.28   &\textbf{13.07}&?\\
         \midrule
        \multicolumn{6}{l}{Our Methods}  \\
        \midrule
         BMHRL& \checkmark & \textbf{4.91} & \textbf{2.23} & 10.80 & \textbf{1012}\\
         BMHRL (weighted)& \checkmark & 4.73 & 2.17 & \textbf{10.92} & 1011\\
         BMHRL (dim. adjustment)& \checkmark & 4.36 & 1.94 & 10.46 & 720\\
         BMHRL (d. f. adjustment)& \checkmark & 4.18 & 1.87 & 10.21 & 816\\
         BMHRL (d. f. = $0$)& \checkmark & x.xx & x.xx & x.xx & 919\\
        \midrule
        \multicolumn{6}{l}{Ablation Study}  \\
        \midrule
         BMH& - & 4.61 & 1.91 & 10.84 & 869\\
         HRL (audio only) & \checkmark & 2.71 & 1.22 & 8.29 & 153\\
         HRL (vision only) & \checkmark & 3.83 & 1.65 & 9.84 & 680\\
         \bottomrule
    \end{tabular}
    }
    \caption{We compare performances of the \ac{BMHRL} iterations (cf.~\Cref{sec:experiments}) with related works on the ActivityNet Captions~\cite{krishna2017densecap} dataset. Best performances are in bold. The highest \ac{METEOR} scores are achieved with \ac{RL} models.}
    \label{tab:model_performances}

    \end{table}
\section{Limitations}
  \noindent\textbf{GT alignment:} The \ac{KL} divergence may only be computed sensibly for sequences of equal length. For a \ac{GT} label of length $n$, any predictions $\hat{y}_{n+1}, ...,\hat{y}_{n+m}$ will not be judged with this models loss function.
  If the prediction contains the \ac{GT}'s content in tokens $\hat{y}_{n+1},...$, the change in \ac{METEOR} score will have no impact on backpropagation.

  \noindent\textbf{Semantics:} While \ac{METEOR} score allows us to account for token permutation and synonyms, it cannot measure the difference of sentence meaning between two sentences. A possible refinement would be to utilise sentence embedding models like SBERT~\cite{reimers-sbert,hier-modular-cap} and measure the change of sentence similarity akin to \Cref{sec:hier-rew}.
\label{sec:limitations}

\section{Conclusion}
\label{sec:conclusion}
We have introduced an attentive \ac{BMHRL} architecture utilising \ac{KL} Divergence to generate video captions that incorporate the long temporal context of input videos and are resilient towards token permutations. The \ac{BLEU} and \ac{METEOR} results indicate the suitability of integrating \ac{HRL} into video captioning models. We have proposed four variants of our original \ac{BMHRL} by adjusting the reward computation (weighted \ac{KL} Divergence), latent feature dimensions, and discount factors. From these three variants, the biased \ac{BMHRL} shows the best overall performances. Moreover, in the ablation study, we showed the complementarity of the audio and video modalities and the effect of their fusion on increasing the \ac{HRL} agent's performance. For future work, it is interesting to explore alternative hierarchical architectures where global context is used to generate both local event-level and video-level rewards~\cite{StreamlinedDenseCap}.
In addition, the use of lightweight, low-resource neural networks such as \textsc{DeepSpectrumLite}~\cite{amiriparian2022deepspectrumlite} can be pursued for real-time feature extraction. Finally, we aim for analysing the effect of fine-tuning our Transformer architecture on broad video captioning datasets such as Charades~\cite{charades} and VATEX~\cite{vatex}.

\begin{acronym}
    \acro{CIDEr}[CIDEr]{Consensus-based Image Description Evaluation}
    \acro{MLE}[MLE]{Maximum likelihood estimation}
    \acro{BLEU}[BLEU]{Bi-Lingual Evaluation Understudy}
    \acro{LSTM}[LSTM]{Long Short-Term Memory}
    \acro{ANN}[ANN]{Artificial Neural Network}
    \acro{CNN}[CNN]{Convolutional Neural Network}
    \acro{GT}[GT]{Ground Truth}
    \acro{CNN}[CNN]{Convolutional Neural Network}
    \acro{DETR}[DETR]{DEtection TRansformer}
    \acro{LN}[LN]{Layer Normalization}
    \acro{I3D}[I3D]{Inflated 3D ConvNet}
    \acro{RL}[RL]{Reinforcement Learning}
    \acro{GloVe}[GloVe]{Global Vectors for Word Representation}
    \acro{FF}[FF]{Feedforward layer}
    \acro{BMT}[BMT]{Bi-Modal Transformer}
    \acro{BMHRL}[BMHRL]{Bi-Modal Hierarchical Reinforcement Learning}
    \acro{GRU}[GRU]{Gated Recurrent Unit}
    \acro{PG}[PG]{Policy Gradient}
    \acro{RNN}[RNN]{Recurrent Neural Network}
    \acro{HRL}[HRL]{Hierarchical Reinforcement Learning}
    \acro{METEOR}[METEOR]{Metric for Evaluation of Translation with Explicit ORdering}
    \acro{mAP}[mAP]{mean Average Precision}
    \acro{NLP}[NLP]{Natural Language Processing}
    \acro{MSE}[MSE]{Mean Squared Error}
    \acro{SOTA}[SOTA]{State Of The Art}
    \acro{DNN}[DNN]{Deep Neural Network}
    \acro{ReLU}[ReLU]{Rectified Linear Unit}
    \acro{AReLU}[AReLU]{Attention Rectified Linear Unit}
    \acro{BN}[BN]{Batch Normalization}
    \acro{KL}[KL]{Kullback-Leibler}
\end{acronym}

{\small
\bibliographystyle{ieee_fullname}
\bibliography{refs}
}

\end{document}